\documentclass[preprint,12pt]{elsarticle}

\usepackage{amssymb}

\usepackage{lineno}

\usepackage{subcaption} 
\usepackage[section]{placeins} 
\usepackage{url} 
\usepackage{pdflscape}

\journal{arXiv}

\begin{document}

\begin{frontmatter}



\title{ToFFi - Toolbox for Frequency-based Fingerprinting of~Brain Signals}


\author[aff-WFAiIS,aff-LNK]{Michał K. Komorowski} 
\ead{michak@is.umk.pl}
\author[aff-LNK]{Krzysztof Rykaczewski}
\author[aff-WFAiIS,aff-LNK]{Tomasz Piotrowski}
\author[aff-Nencki,aff-Montreal]{Katarzyna Jurewicz}
\author[aff-Kajetany,aff-Nencki]{Jakub Wojciechowski}
\author[aff-Dundee]{Anne Keitel}
\author[aff-Human]{\\Joanna Dreszer}
\author[aff-WFAiIS,aff-LNK]{Włodzisław Duch}

\address[aff-WFAiIS]{Department of Informatics, Faculty of Physics, Astronomy, and Informatics, Nicolaus Copernicus University, Toruń, Poland}
\address[aff-LNK]{Neurocognitive Laboratory, Centre for Modern Interdisciplinary Technologies, Nicolaus Copernicus University, Toruń, Poland}
\address[aff-Nencki]{Nencki Institute of Experimental Biology, Polish Academy of Sciences, Warsaw, Poland}
\address[aff-Montreal]{Department of Neurosciences, Faculté de médecine, Université de Montréal, Montréal, QC, Canada}
\address[aff-Kajetany]{Bioimaging Research Center, Institute of Physiology and Pathology of Hearing, Kajetany, Poland}
\address[aff-Dundee]{Psychology, University of Dundee, Scrymgeour Building, Dundee, United Kingdom}
\address[aff-Human]{Faculty of Philosophy and Social Sciences, Institute of Psychology, Nicolaus Copernicus University in Toruń, Poland}

\begin{abstract}
 
Spectral fingerprints (SFs) are unique power spectra signatures of human brain regions of interest (ROIs, Keitel \& Gross, 2016). SFs allow for accurate ROI identification and can serve as biomarkers of differences exhibited by non-neurotypical groups. At present, there are no open-source, versatile tools to calculate spectral fingerprints. We have filled this gap by creating a modular, highly-configurable MATLAB Toolbox for Frequency-based Fingerprinting (ToFFi). It can transform MEG/EEG signals into unique spectral representations using ROIs provided by anatomical (AAL, Desikan-Killiany), functional (Schaefer), or other custom volumetric brain parcellations. Toolbox design supports reproducibility and parallel computations. 

\end{abstract}


\end{frontmatter}




\section{Introduction}
\label{introduction}

\begin{figure}[htbp]
	\centering
	\includegraphics[width=0.85\linewidth]{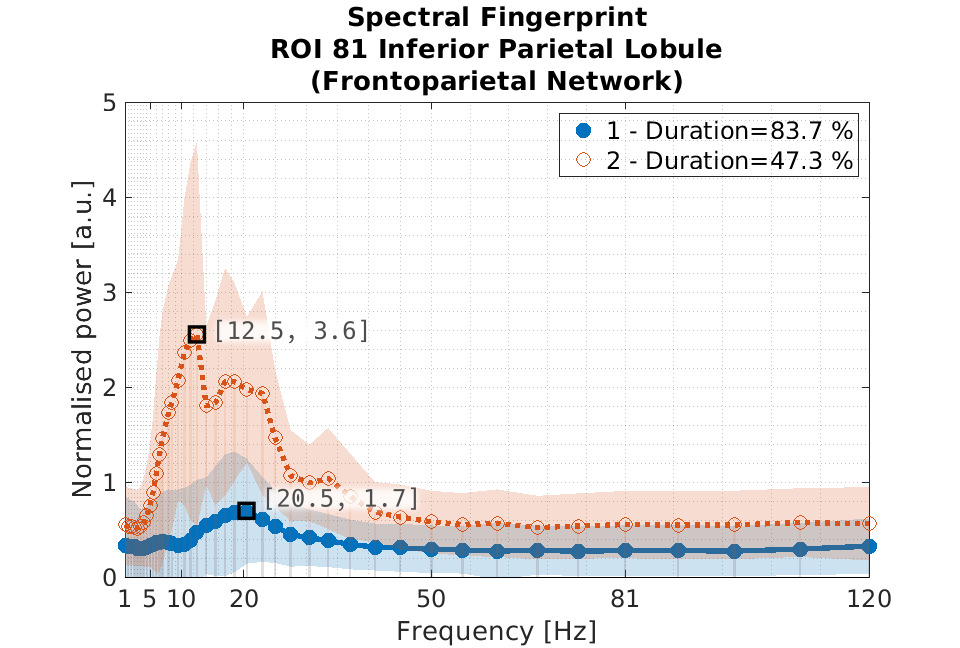}	
	\caption{A spectral fingerprint of the inferior parietal lobule. For this particular region, it consists of two spectral modes. It is formed by clustering power spectra segments (normalized, i.e. spectral power in comparison to the whole brain) first on the individual subjects level and then clustered again on the group level. Each mode corresponds to one of the centroids found by the clustering algorithm. Shaded regions depict the standard deviation ($1\sigma$) estimated from the covariance matrix of the Gaussian Mixture Model component corresponding to the given spectral mode. The first mode peaks at 12.5 Hz, and the second mode peaks at 20.5 Hz. The frequency axis resolution can be set to logarithmic to optimize spectral analysis resolution of lower frequencies. Duration is shown as a percentage of time segments in which each spectral mode was present on average during recording.
	}
	\label{fig:SF-ex}
\end{figure}

Brain dynamics and brain oscillations are among the most important topics in neuroscience. Different methods proved to be useful for studying robust whole-brain, as well as regionally-specific patterns of activity, called brain fingerprints. They can serve as signatures for mental states during task execution or rest \cite{bola_dynamic_2015}, \cite{krienen_reconfigurable_2014}, \cite{ciric_contextual_2017}, \cite{keynan_limbic_2016}. Frequency of oscillations turned out to be one of the key features in many studies describing particular regions of interest (ROIs) \cite{siegel_spectral_2012}, \cite{mellem_intrinsic_2017}, \cite{keitel_individual_2016} and \mbox{large-scale} brain networks \cite{mahjoory_frequency_2020}, \cite{samogin_shared_2019}, \cite{marino_neuronal_2019}, \cite{hacker_frequency-specific_2017}, \cite{rosanova_natural_2009}. Spectral fingerprints (Fig.~\ref{fig:SF-ex}) play a role as biomarkers that are sufficiently specific to permit the successful identification of brain regions using their spectral characteristics. Moreover, spectral profiles’ peaks that correspond to the natural frequencies of ROIs \cite{rosanova_natural_2009}, \cite{ferrarelli_reduced_2012}, are consistently modulated by specific tasks, neurological or mental disorders. They can be generalized across groups of participants \cite{lubinus_data-driven_2021}, \cite{keitel_individual_2016}. In this paper, we introduce a novel implementation of the Spectral Fingerprinting technique, in a highly configurable MATLAB toolbox.

\section{Problems and Background}
\label{problems-and-background}



There are many open software packages available to analyze neural data. The Fieldtrip Toolbox\footnote{\url{https://www.fieldtriptoolbox.org/} ; accessed: 14.10.2021} \cite{oostenveld_fieldtrip:_2010} was designed to perform analysis both on sensor and source level of EEG/MEG/iEEG/NIRS data. EEGLAB\footnote{\url{https://eeglab.org/} ; accessed: 14.10.2021} \cite{delorme_eeglab_2004} helps with processing continuous and event-related electrophysiological data implementing many analytic methods (ICA, time/frequency analysis, artifact rejection, event-related statistics, microstates analysis) and several useful routines for visualization. To simulate brain dynamics, perform connectivity analyses, and solve forward/inverse problems, the supFunSim\footnote{\url{https://github.com/nikadon/supFunSim} ; accessed: 14.10.2021} toolbox \cite{rykaczewski_supfunsim_2021} and the Virtual Brain\footnote{\url{https://www.thevirtualbrain.org/} ; accessed: 14.10.2021} system \cite{sanz_leon_virtual_2013} are among suitable choices. However, there is no open software for analyzing spectral fingerprints, and our work attempts to fill this gap. We designed the Toolbox for Frequency-based Fingerprinting (ToFFi, \url{https://github.com/micholeodon/ToFFi_Toolbox})
for analysis of MEG, EEG, and other multichannel data. Users can configure many parameters for each stage of processing, including the selection of the brain parcellation, and decide which of them will run in parallel (cluster computations are supported). Results of the calculations are reproducible thanks to the implemented control using pseudo-random number generators and visualization scripts.

\section{Software Framework }
\label{software-framework}

\subsection{Software Architecture}
\label{software-architecture}

\begin{figure}[htbp]
	\centering
	\includegraphics[width=1.0\linewidth]{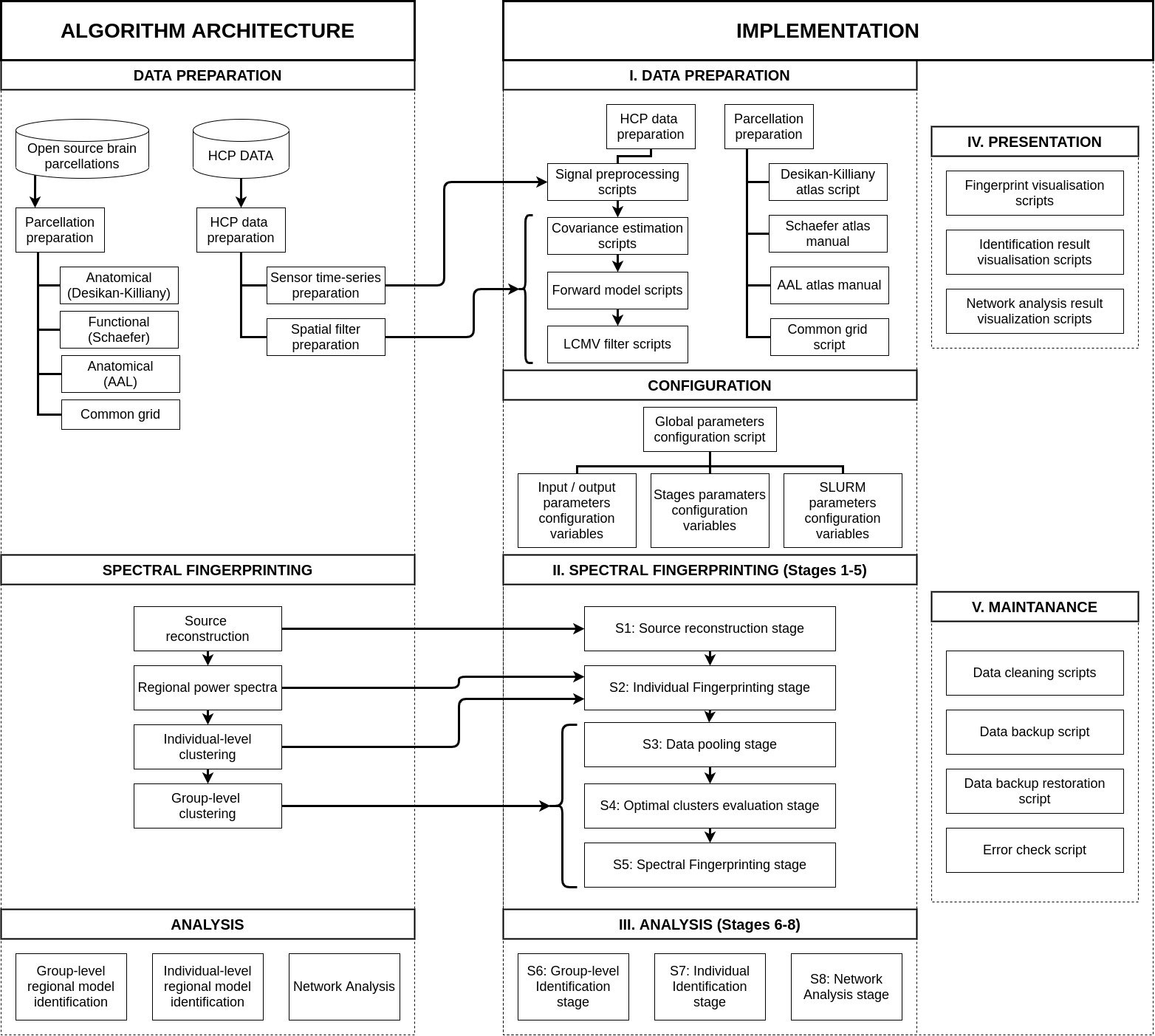}	
	\caption{ToFFi toolbox architecture.}
	\label{fig:architecture}
\end{figure}
\begin{figure}[htbp]
	\centering
	\includegraphics[width=1.0\linewidth]{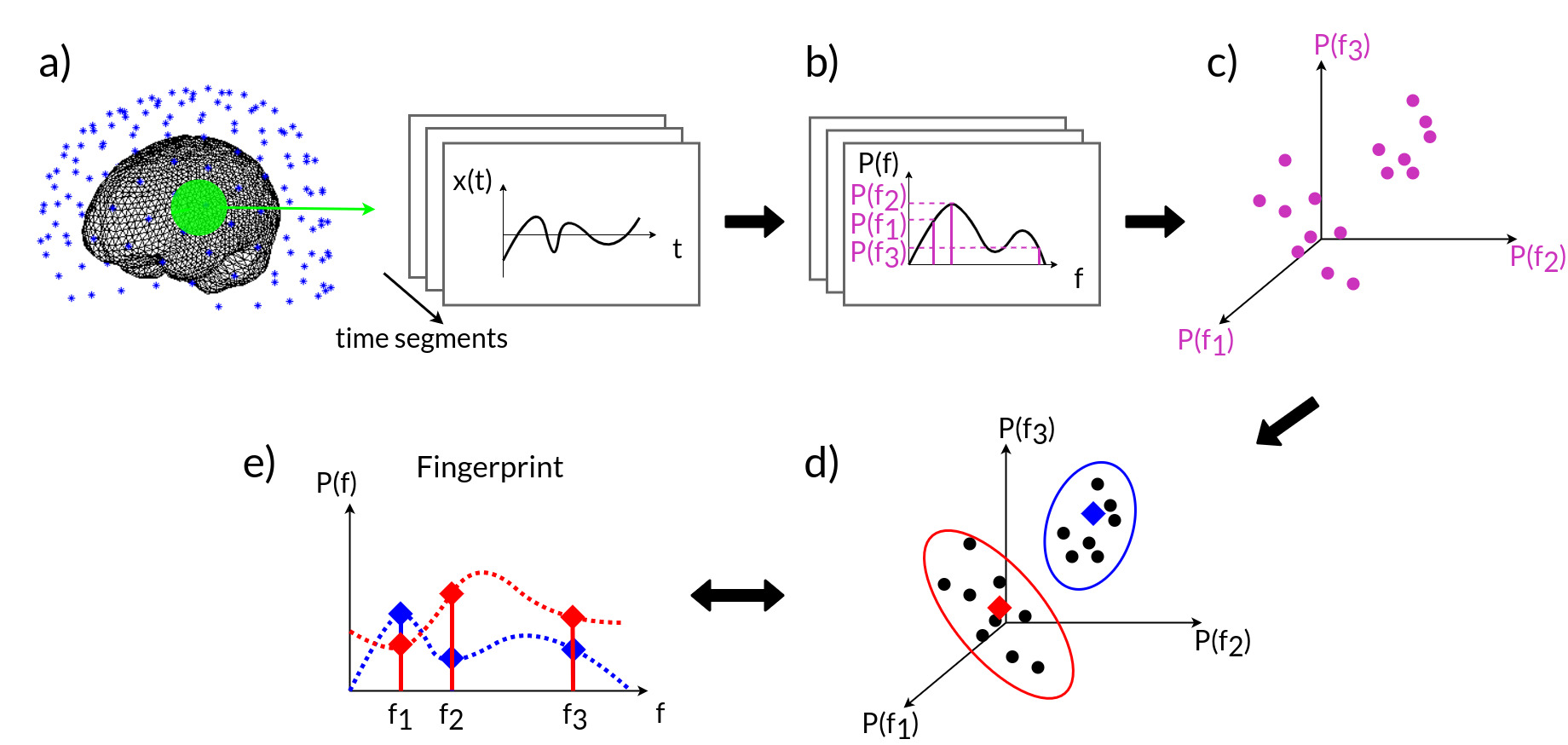}
	\caption{Illustration of the Spectral Fingerprinting algorithm \cite{keitel_individual_2016}. a) Mean-field, electrical activity of the brain is recorded via an array of sensors (blue dots). A beamformer \cite{sekihara_adaptive_2008} solves so-called \textit{inverse problem}, thus enabling reconstruction of the source activity in selected voxels constituting chosen brain regions of interest (green area). Source-level time series are cut into segments of equal length. b) Power within each segment is estimated for a given set of frequencies of interest (here three frequencies, $f_1$, $f_2$, $f_3$, are shown for readability) and then averaged across locations inside the region of interest. c) Power spectrum of each segment is represented as a point in a $n$-dimensional frequency space, where power of the selected frequencies provides coordinates of the point. d) Segments are clustered together. Each centroid is equivalent to one \textit{spectral mode} of a given brain region and depicted as an interpolant curve spanned between the frequencies of interest in the resulting fingerprint (interpolation type is arbitrary, as it serves visualization purpose only).}
	\label{fig:fingerprinting}
\end{figure}
\begin{figure}[htbp]
	\centering
	\includegraphics[width=1\linewidth]{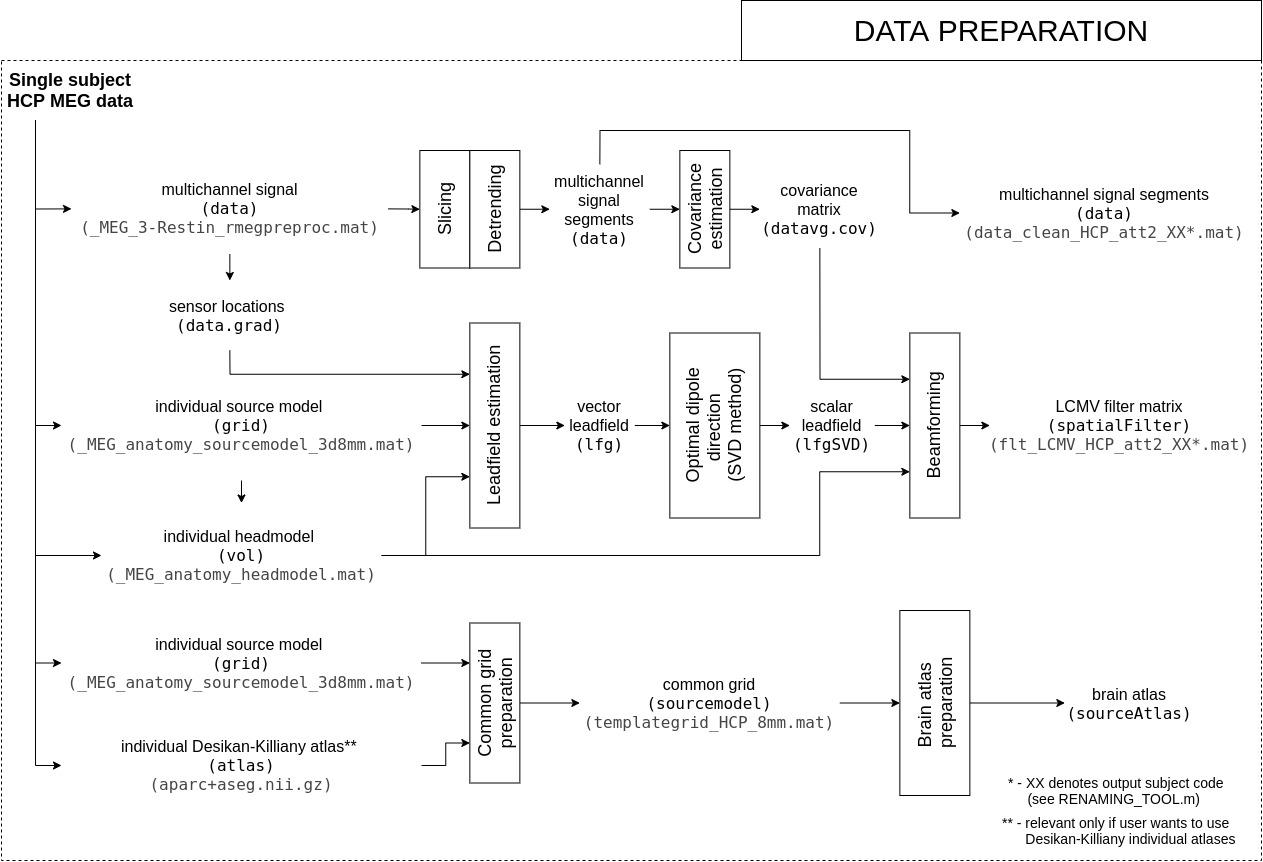}
	\caption{Diagram of how the data are processed by the DATA PREPARATION component (see Fig. \ref{fig:architecture}). Each depicted piece of data is endowed with its name, MATLAB workspace variable name (middle parentheses), and a file name (bottom parentheses) if it is read or written to the disk during processing.}
	\label{fig:data-flow-data-prep}
\end{figure}
\begin{figure}[htbp]
	\centering
	\includegraphics[width=1\linewidth]{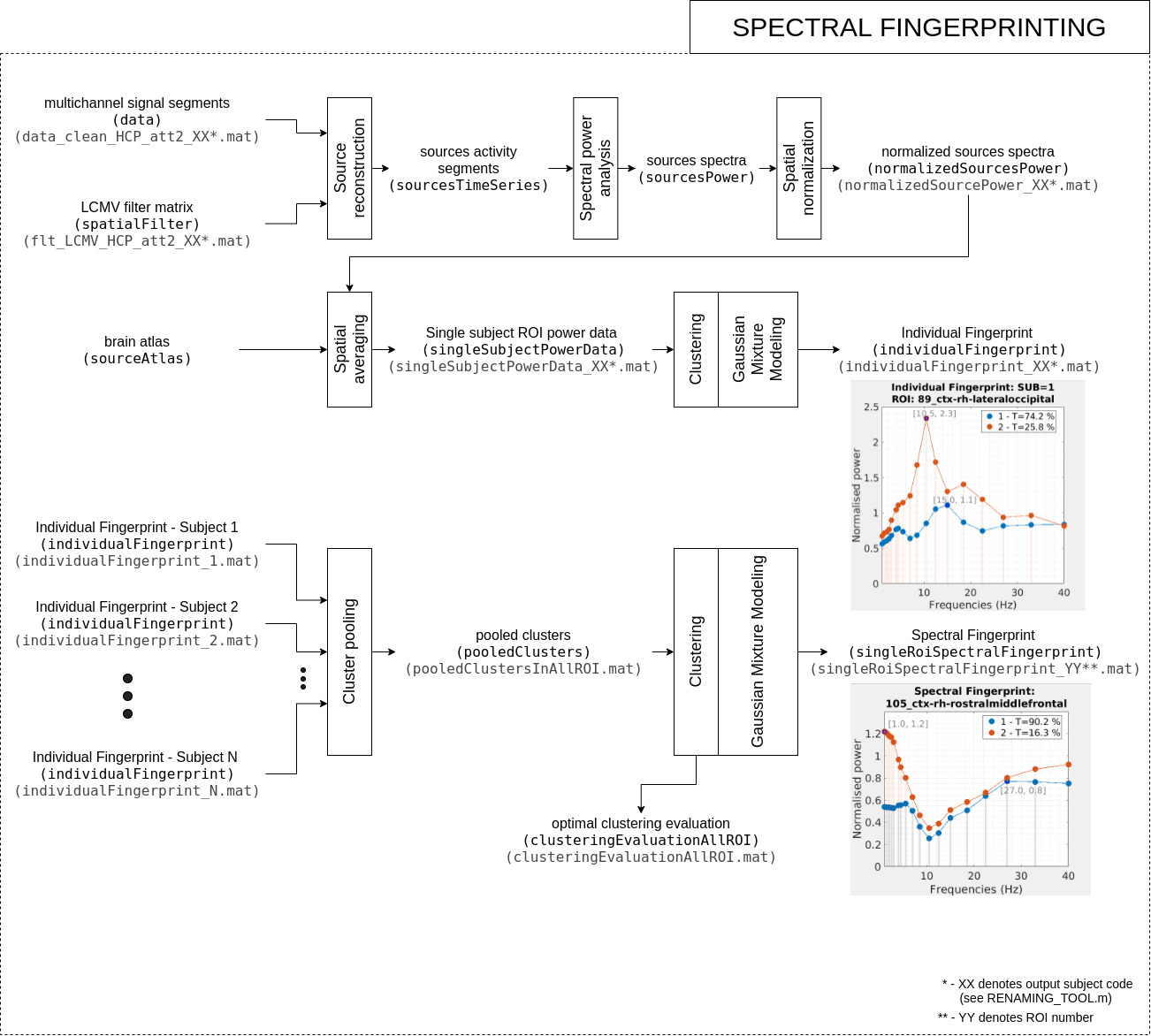}
	\caption{Diagram of how the data are processed by the SPECTRAL FINGERPRINTING component (see Fig. \ref{fig:architecture}). Each depicted piece of data is endowed with its name, MATLAB workspace variable name (middle parentheses), and a file name (bottom parentheses) if it is read or written to the disk during processing.}
	\label{fig:data-flow-sf}
\end{figure}
\begin{figure}[htbp]
	\centering
	\includegraphics[width=1\linewidth]{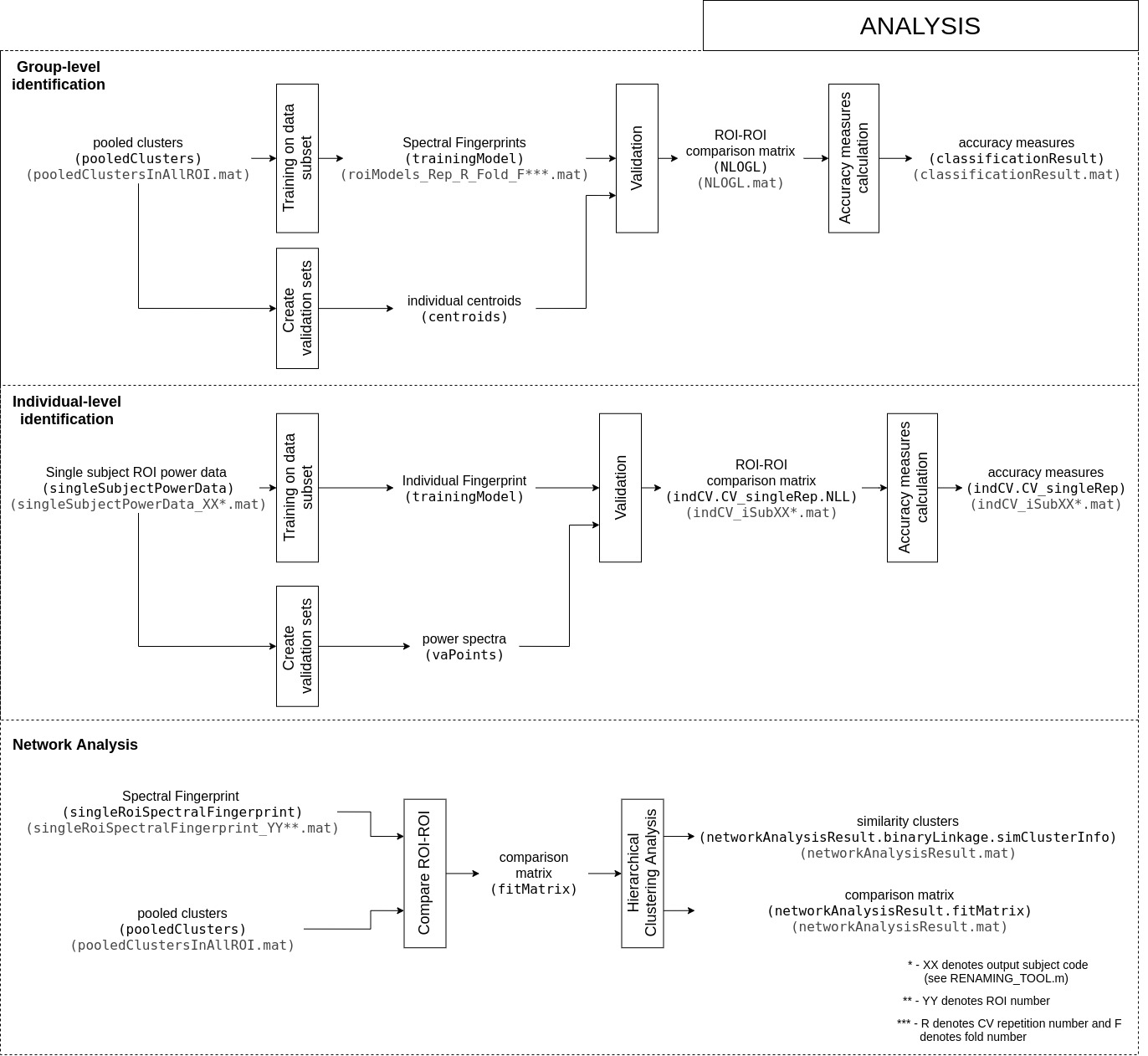}
	\caption{Diagram of how the data are processed by the ANALYSIS component (see Fig. \ref{fig:architecture}). Each depicted piece of data is endowed with its name, MATLAB workspace variable name (middle parentheses), and a file name (bottom parentheses) if it is read or written to the disk during processing.}
	\label{fig:data-flow-analysis}
\end{figure}

ToFFi is a modular piece of software that consists of five components: I.~Data Preparation, II.~Spectral Fingerprinting, III.~Analysis, IV.~Presentation, and V.~Maintenance (Fig. \ref{fig:architecture}). The Data Preparation module (I) is responsible for arranging sensor time series signals, spatial filters, and brain parcellation data, for processing by the second step routines. The Spectral Fingerprinting (II) module transforms MEG/EEG multichannel array of signals, through a series of five stages, into spatially localized power spectrum-driven representations called spectral fingerprints (Fig. \ref{fig:fingerprinting}). Fourier Transform, source reconstruction (beamforming), and Gaussian Mixture Modeling algorithms are used to compute spectral fingerprints both at the individual and the group level. The third component (III) consists of additional routines that can analyze particular output files from component II. Currently, we have implemented group-level brain regions identification, individual-level brain regions identification, and regional clustering (network analysis) - all based on the concept of modeling brain activity as spectral fingerprints. The Presentation component (IV) is a collection of auxiliary scripts used to visualize particular results of performed computations for easier interpretation. Maintenance routines (V) are used to automate some parts of the workflow, e.g.: manage configuration files, manage output data files, etc.
A more detailed description of how the data are transformed can be found in \ref{apdx-toffi-toolbox-manual} (6. METHODS), Fig. \ref{fig:data-flow-data-prep}, Fig. \ref{fig:data-flow-sf}, and Fig. \ref{fig:data-flow-analysis}, which summarize the whole pipeline.

\subsection{Software Functionalities}
\label{software-functionalities}



Spectral Fingerprinting can be performed on multichannel time series data (e.g. MEG, EEG) of arbitrary size, with any sampling-frequency adjusted to the desired frequency resolution, acquired from a single multiple subjects, and divided into segments of selected, equal duration (e.g. 1000 ms). These segments may contain non-overlapping pieces of a continuous recording (e.g. resting-state) or trials with brain responses for several repetitions of the same experimental condition (event-related paradigm). For individual-level analysis, the toolbox offers a reconstruction of voxel-wise time series power spectra with beamforming using precomputed spatial filters (e.g. LCMV, \cite{van_veen_localization_1997}, \cite{sekihara_adaptive_2008}) and multichannel empirical sensor signals or artificial white Gaussian noise signals. Power spectra of ROIs can be estimated both at individual and group level, using different brain parcellations (anatomical: AAL \cite{tzourio-mazoyer_automated_2002}, Desikan-Killiany \cite{desikan_automated_2006}; functional: Schaefer  \cite{schaefer_local-global_2018}), and optionally normalized. These spectra can be clustered (currently, only the k-means algorithm is implemented) with arbitrary distance metric and subsequently modeled as a regularized Gaussian mixture of regional spectra with a fixed or optimal number of clusters to construct group-level fingerprints. The user can also estimate the accuracy of identification of brain regions from their spectral fingerprints using cross-validation. Hierarchical clustering of spectral fingerprints (network analysis) was also implemented. The scope of selected brain regions of interest (ROIs) and set of subjects of choice can be limited if desired. For the majority of stages, one can perform computations in parallel on a single computer with multiple cores, or on a grid of multiple-core machines orchestrated via workload manager (currently, only SLURM manager is supported). Interpretation of the outputs of the software components II and III are supported with visualization routines. For reproducibility, data maintenance routines and pseudo-random generator control are implemented as well.

\section{Implementation and Empirical Results}
\label{implementation-and-empirical-results}

\subsection{Implementation}
\label{implementation}

The toolbox can be operated under Linux, macOS, and Windows systems. Maintenance scripts (V) are coded mostly in Bash, which is accessible both for Linux (as default) and Windows (using Cygwin, cmder, or other shell emulator). All calculations are carried out entirely in MATLAB (version R2021a or newer is recommended) with Signal Processing Toolbox, Statistics and Machine Learning Toolbox, Parallel Computing Toolbox, and open-source Fieldtrip Toolbox (version 20210816 or newer is preferred, \cite{oostenveld_fieldtrip:_2010}). Additionally, \texttt{vline.m} and \texttt{hline.m} functions\footnote{\url{https://www.mathworks.com/matlabcentral/fileexchange/1039-hline-and-vline} ; accessed: 14.10.2021} by Brandon Kuczenski are used for plotting, and \texttt{HZmvntest.m} function\footnote{\url{https://www.mathworks.com/matlabcentral/fileexchange/17931-hzmvntest} ; accessed: 14.10.2021} by Antonio Trujillo-Ortiz for multivariate normality testing. If the user wishes to enable cluster computations, the toolbox is prepared to work in coordination with the SLURM workload manager\footnote{\url{https://slurm.schedmd.com/} ; accessed: 14.10.2021}. To the best of our knowledge, there is no other software for Spectral Fingerprinting available, apart from the illustrative beta-version script referenced by the authors of \cite{keitel_individual_2016}.

\subsection{Empirical results}
\label{empirical-results}



Keitel and Gross \cite{keitel_individual_2016} showed that rendering regional brain activity as a combination of spectra via Spectral Fingerprinting allows for the identification of ROIs with high accuracy. They noticed that clustering of the brain areas according to the similarity of spectral profiles shows patterns similar to macroscale organization of the human brain cortex. Auditory spectral profiles turned out to be modulated during auditory processing. Lubinus and colleagues \cite{lubinus_data-driven_2021} have discovered that visual deprivation is reflected in the modulation of spectral fingerprints, indicating possible correspondence with the structural and functional adaptation of the human brain. Likewise, Mellem with collaborators \cite{mellem_intrinsic_2017} demonstrated via a similar method that there is a mix of lower and higher frequency peaks across the brain and it does not follow a simple lower order-higher order processing hierarchy. 

\section{Illustrative Example}
\label{illustrative-example}



Please consult the following parts of \ref{apdx-toffi-toolbox-manual} to run Illustrative Example smoothly: \textit{Chapter 2. Conventions} - to learn notation used throughout the documentation; \textit{Chapter 3. Installation} - to set up a computational environment properly, \textit{Sections 5.3} and \textit{5.4} - to get the input data.

After installation, one is advised to follow the instructions in \textit{Chapter~4.~Illustrative Example} in \ref{apdx-toffi-toolbox-manual} to complete the illustrative example using the Human Connectome Project MEG dataset (HCP Reference Manual, \cite{van_essen_wu-minn_2013}). We have selected N=10 subjects with the MEG resting-state cleaned signal acquired via a 248 channel array in three subsequent runs, approximately 3 min each. 

Spectral Fingerprinting routines were configured to optimize the frequency resolution for the lower frequencies, thus accounting for the $1/f$ power trend present in the typical electrophysiological activity of the human brain. The proper number of clusters to be constructed was estimated using the Silhouette optimality criterion. Choosing cosine dissimilarity as a distance measure helped to compose frequency clusters of power spectra similar in shape, diminishing the influence of the power spectra amplitude. To speed up computations, the number of CPU cores was set to two.

\begin{figure}[htbp]
	\centering
	\includegraphics[width=0.7\linewidth]{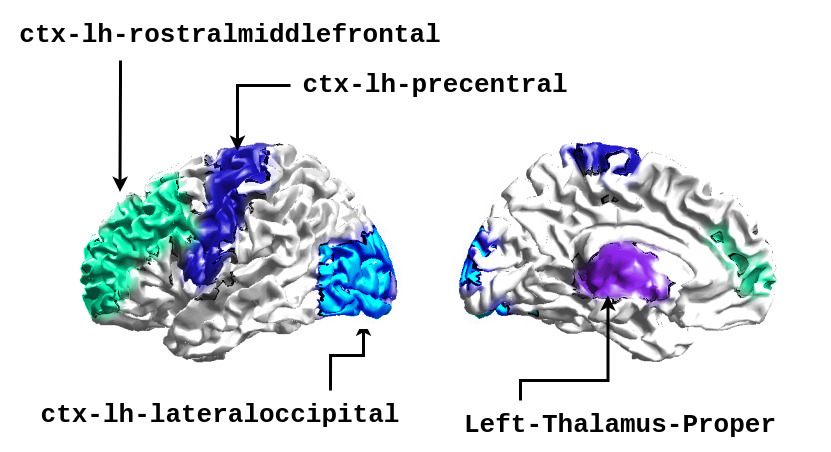}	
	\caption{Brain regions chosen from the Desikan-Killiany atlas for the purpose of the illustrative example. Here only the left counterpart is depicted, whereas right hemisphere homologues were chosen as well.}
	\label{fig:ill-roi}
\end{figure}

\begin{figure}[htbp]
	\begin{subfigure}{.49\textwidth}
		\centering
		\includegraphics[width=1\linewidth]{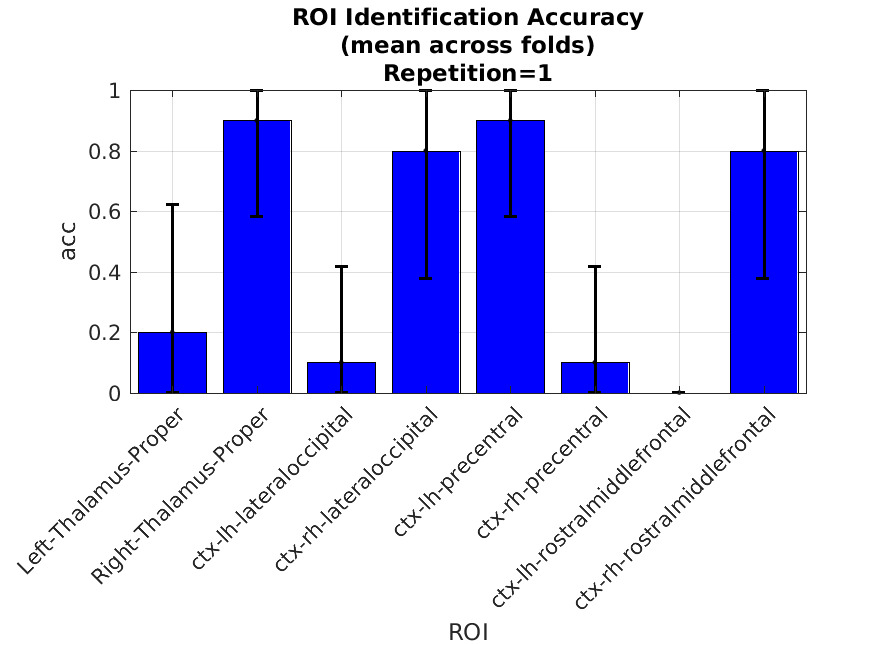}
		\caption{}
		\label{fig:s6-acc_folds}
	\end{subfigure}
	\hfill
	\begin{subfigure}{.49\textwidth}
		\centering
		\includegraphics[width=1\linewidth]{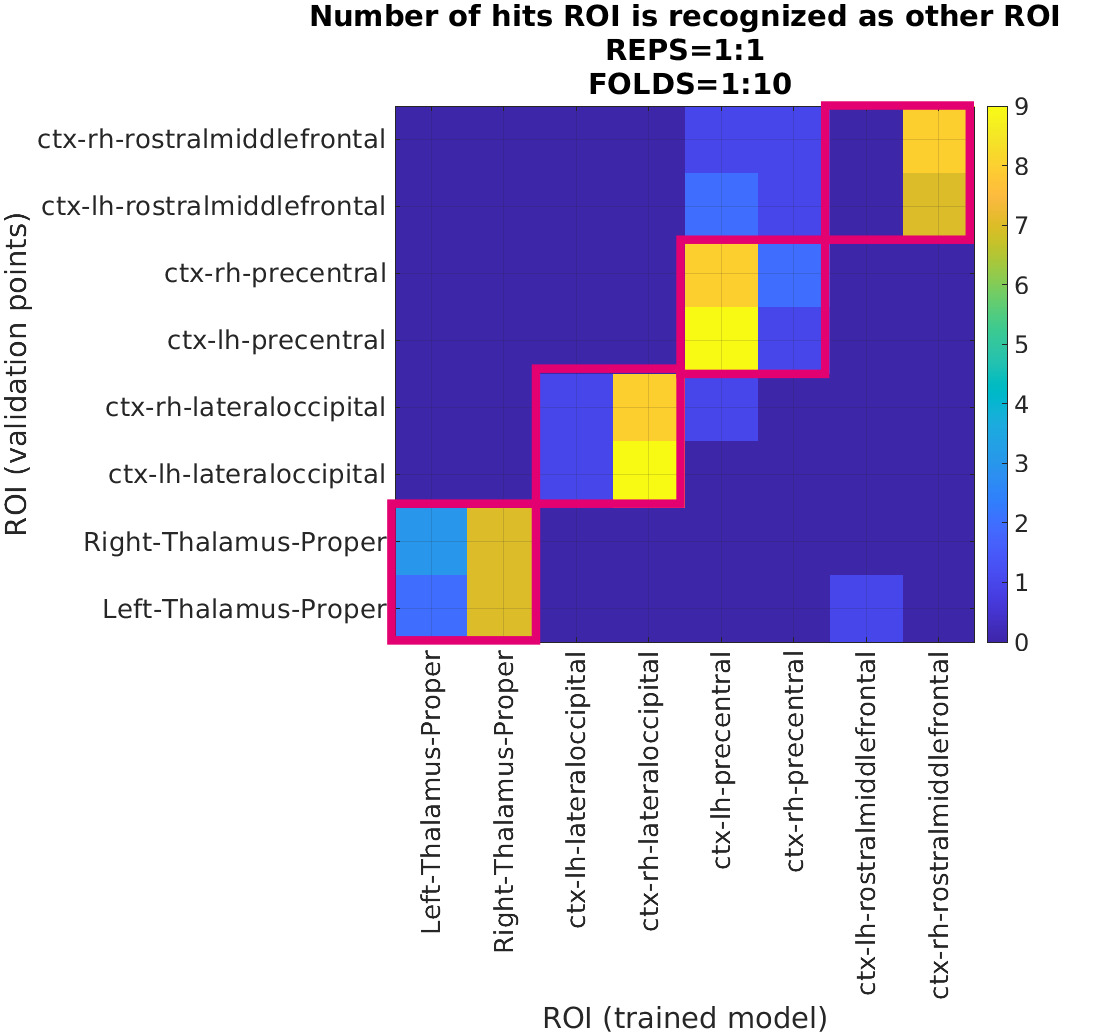}
		\caption{}
		\label{fig:s6-hit-matrix}
	\end{subfigure}
	\caption{Group-level identification accuracy: a) bar plot showing the average identification accuracy across cross-validation iterations (leave-one-out), b) confusion matrix showing in each row a distribution of "votes" for each ROI. Each ROI was tested ten times (model trained on nine subjects versus one validation subject). For ideal identification, this matrix would have a value of 10 for the diagonal elements and zeros elsewhere. Confusion happens mostly between homologue areas (2x2 red boxes). Left hemisphere ROIs are recognized as the right hemisphere homologue areas. }
	\label{fig:illustrative-s6}
\end{figure}

Group-level fingerprints in the 1--40\,Hz frequency interval from 8 distant regions of the human brain (ROIs; Fig. \ref{fig:ill-roi}) were found (Fig. \ref{fig:illustrative-SF}). The similarity of fingerprints was assessed (Fig. \ref{fig:illustrative-network-analysis}), together with the accuracy of how well one can identify them (Fig. \ref{fig:illustrative-s6}). 

The proposed method allowed for discrimination between different modes of operation for a range of brain areas. Dominant and supportive group-level oscillation profiles were recognized and separated. Functional similarity between homologue areas was confirmed using hierarchical clustering analysis. Recognition of the brain areas based on their spectral fingerprints turned out to be challenging among homologue areas due to their functional similarity, yet remaining informative.

\section{Conclusions}
\label{conclusions}

Spectral Fingerprinting allows for the discovery of meaningful oscillatory patterns from electrophysiological time series that can show task-induced modulations or serve as a signature of the brain’s regional activity 
in the particular parcellation. Our novel Toolbox for Frequency-based Fingerprinting (ToFFi) provides researchers with a modular, highly configurable tool for computing regional source-reconstructed power spectra, finding optimal prototypes common for a group of subjects via individual- and group-level clustering algorithms, together with testing their properties using additional analytical routines. The efficiency is boosted with parallel computation support, and reproducibility is controlled with pseudo-random number generator parameters. An in-depth understanding of the underlying algorithms is facilitated by the function reference (\ref{apdx-functions-reference}) and the toolbox manual (\ref{apdx-toffi-toolbox-manual}). Presented software is compatible with various modern tools used by the neuroscientific community and allows easy adaptation of its modular structure to specific tasks.  

\section*{Acknowledgments}
\label{akcnowledgements}


Funding: This work has been supported by the National Science Centre, Poland, grant UMO-2016/20/W/NZ4/00354. Data were provided [in part] by the Human Connectome Project, WU-Minn Consortium (Principal Investigators: David Van Essen and Kamil Ugurbil; 1U54MH091657) funded by the 16 NIH Institutes and Centers that support the NIH Blueprint for Neuroscience Research; and by the McDonnell Center for Systems Neuroscience at Washington University. Calculations were carried out at the Tricity Academic Supercomputer \& Network Center in Gdańsk. We thank professor Joachim Gross for his generous technical support, Ewa Ratajczak, and Bartosz Kochański for helping out with testing and debugging the software.


\newpage

\section*{Figures}
\label{apdx-figures}

\FloatBarrier
\newpage
%

\FloatBarrier
\newpage
%

\FloatBarrier
\newpage
%

\FloatBarrier
\newpage
%

\FloatBarrier
\newpage
%

\FloatBarrier
\newpage
%

\FloatBarrier
\newpage
\begin{landscape}
\begin{figure}[htbp]
	\centering
	\begin{subfigure}{.34\textwidth}
		\centering
		\includegraphics[width=1\linewidth]{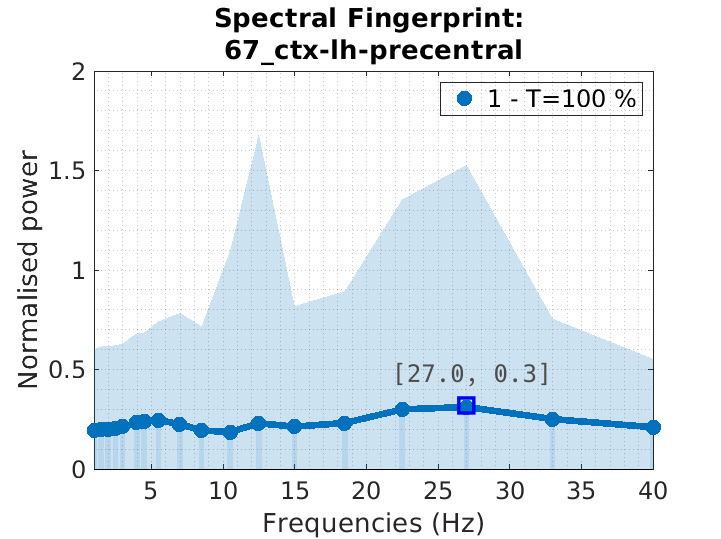}
		\label{fig:SF_ex1}
	\end{subfigure}
	\begin{subfigure}{.34\textwidth}
		\centering
		\includegraphics[width=1\linewidth]{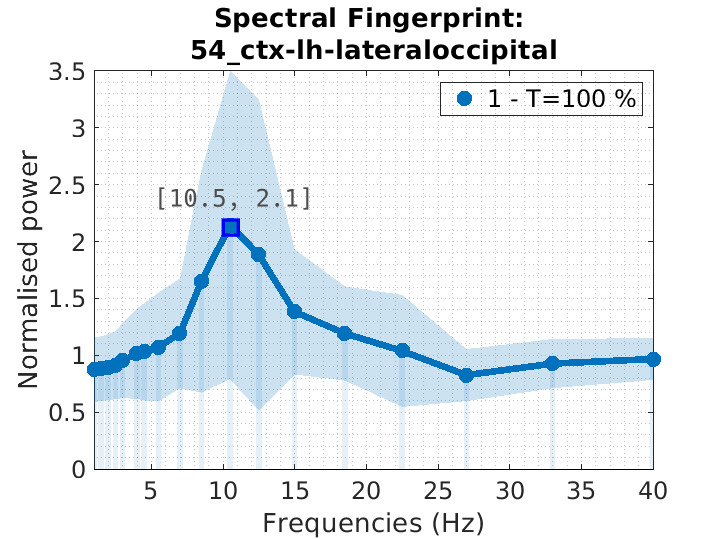}
		\label{fig:SF_ex2}
	\end{subfigure}
	\begin{subfigure}{.34\textwidth}
		\centering
		\includegraphics[width=1\linewidth]{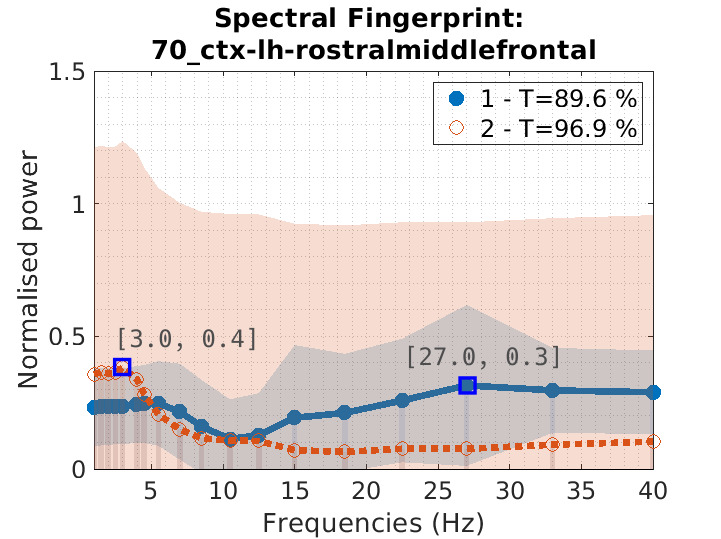}
		\label{fig:SF_ex3}
	\end{subfigure}
	\begin{subfigure}{.34\textwidth}
		\centering
		\includegraphics[width=1\linewidth]{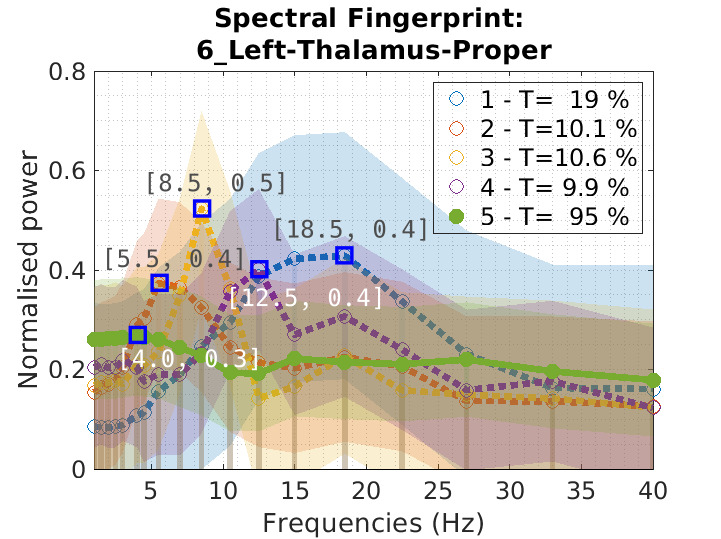}
		\label{fig:SF_ex4}
	\end{subfigure}
\\
	\begin{subfigure}{.34\textwidth}
		\centering
		\includegraphics[width=1\linewidth]{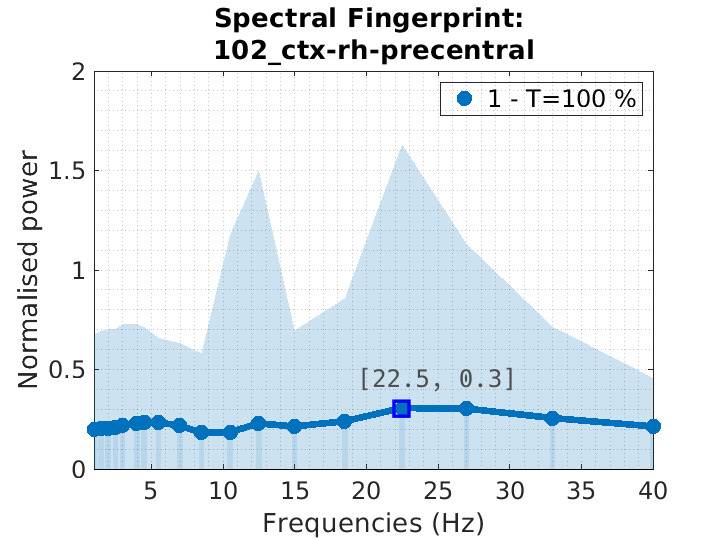}
		\label{fig:SF_ex5}
	\end{subfigure}	
	\begin{subfigure}{.34\textwidth}
		\centering
		\includegraphics[width=1\linewidth]{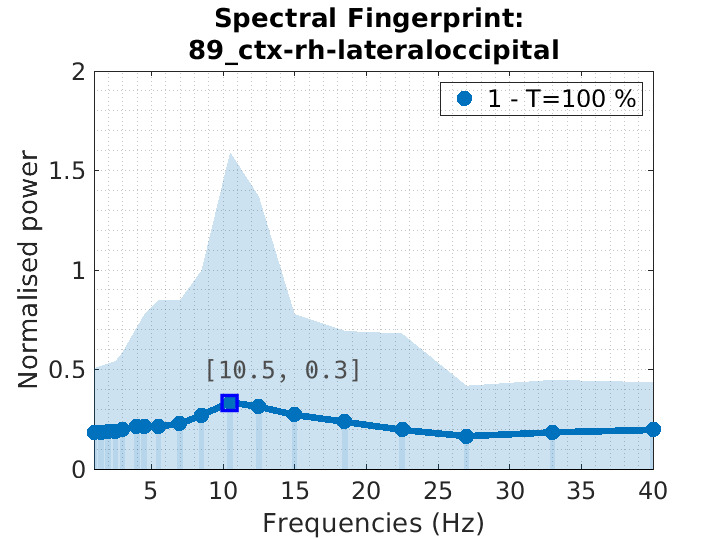}
		\label{fig:SF_ex6}
	\end{subfigure}
	\begin{subfigure}{.34\textwidth}
		\centering
		\includegraphics[width=1\linewidth]{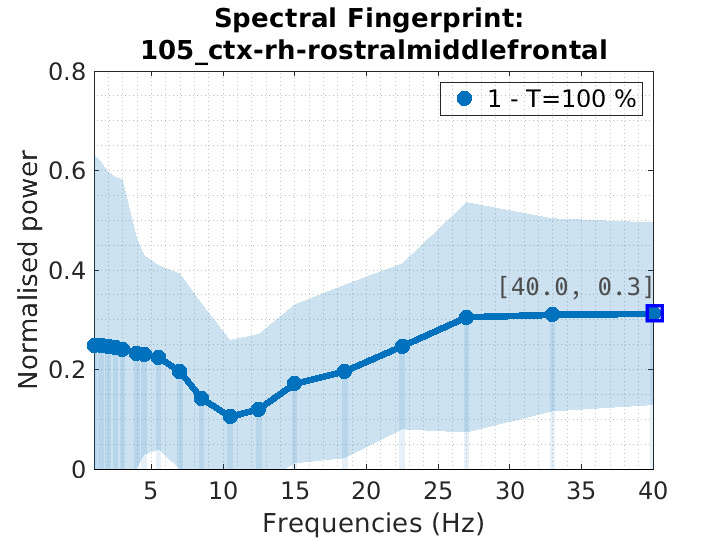}
		\label{fig:SF_ex7}
	\end{subfigure}
	\begin{subfigure}{.34\textwidth}
		\centering
		\includegraphics[width=1\linewidth]{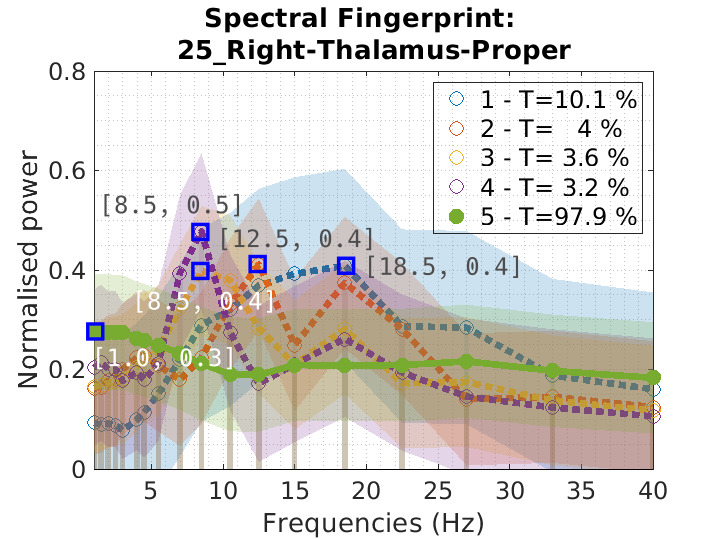}
		\label{fig:SF_ex8}
	\end{subfigure}
	
	\caption{(in color) Resting-state spectral fingerprints for Desikan-Killiany atlas in the 1--40\,Hz frequency interval. Each column shows two homologue brain areas. Legends show the corresponding duration of each spectral mode (i.e., the percentage of trials in which each spectrum was present on average during recording) and whether the mode was present for at least five subjects (filled dot) or not (empty dot). The frequency axis was configured to be logarithmic in order to optimize the lower frequencies resolution. Y-axis depicts the power normalized in relation to the average spectrum of the whole brain. Shaded regions depict the standard deviation ($1\sigma$) of the corresponding spectral mode. For $i$-th of total $F$ frequencies of interest, standard deviation was estimated as $\sqrt{\Sigma_{i,i}}$, where $\Sigma_{i,i}$ is the $i$-th diagonal entry of the covariance matrix of the Gaussian Mixture Model component corresponding to the given spectral mode. Standard deviations have relatively large values due to the small number of subjects used in the illustrative example. Homologue areas have very similar fingerprints.}
	\label{fig:illustrative-SF}
\end{figure}

\FloatBarrier

\end{landscape}
\newpage
%

\begin{figure}[htbp]
	\begin{subfigure}{.32\textwidth}
		\centering
		\includegraphics[width=0.9\linewidth]{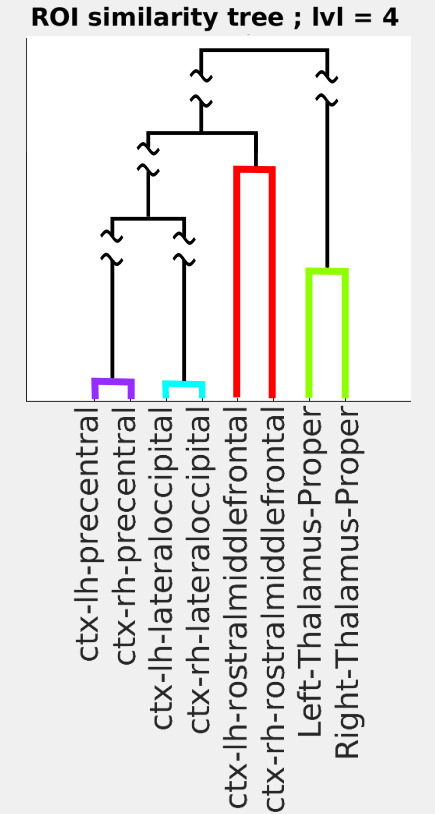}
		\label{fig:tree}
	\end{subfigure}
	\begin{subfigure}{.32\textwidth}
		\centering
		\includegraphics[width=0.9\linewidth]{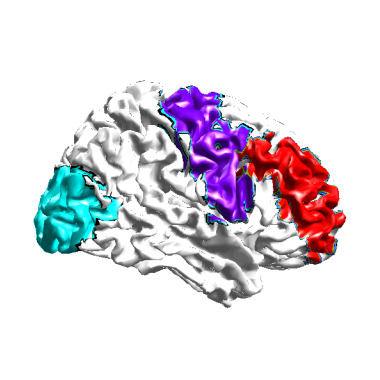}
		\label{fig:both}
	\end{subfigure}
	\begin{subfigure}{.32\textwidth}
		\centering
		\includegraphics[width=0.9\linewidth]{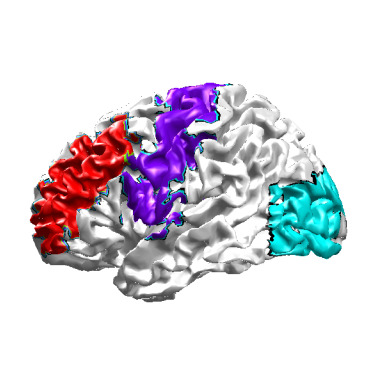}
		\label{fig:left-out}
	\end{subfigure}
	\begin{subfigure}{.32\textwidth}
		\centering
		\includegraphics[width=0.9\linewidth]{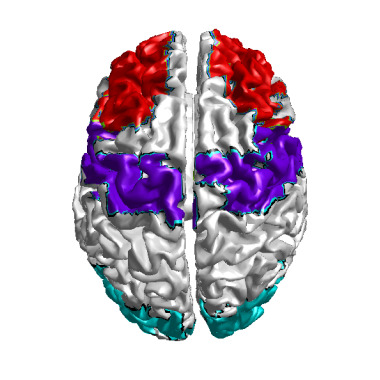}
		\label{fig:right-out}
	\end{subfigure}
	\begin{subfigure}{.32\textwidth}
		\centering
		\includegraphics[width=0.9\linewidth]{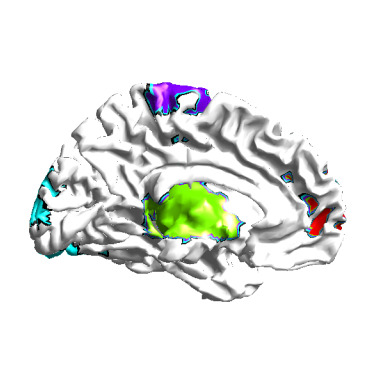}
		\label{fig:left-in}
	\end{subfigure}	
	\begin{subfigure}{.32\textwidth}
		\centering
		\includegraphics[width=0.9\linewidth]{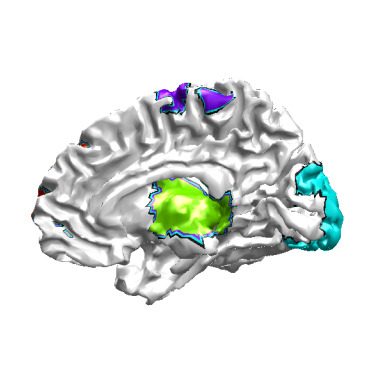}
		\label{fig:right-in}
	\end{subfigure}	
	\caption{Result of the hierarchical agglomerative clustering of the spectral fingerprints presented in Fig. \ref{fig:illustrative-SF}. Homologue areas were automatically matched together according to the similarity of their fingerprints. The similarity tree has disproportionately long branches that were broken for clarity (waved lines).}
	\label{fig:illustrative-network-analysis}
\end{figure}

\FloatBarrier
\newpage
%

\FloatBarrier
\newpage

\FloatBarrier

\appendix

\section{ToFFi Toolbox Manual}
\label{apdx-toffi-toolbox-manual}

Link: \url{https://github.com/micholeodon/ToFFi_Toolbox/tree/master/ToFFi_Toolbox-20211013/docs/ToFFi_Manual.pdf}

\section{Functions reference}
\label{apdx-functions-reference}

Functions reference documents most important M-File Functions of the ToFFi Toolbox. 

It can be accessed \textbf{after} downloading/cloning the ToFFi Toolbox repository (\url{https://github.com/micholeodon/ToFFi_Toolbox}).

Functions reference can be found here: \\\texttt{ToFFi\_Toolbox-YYYYMMDD/docs/FUNCTIONS\_REFERENCE.html}, where \texttt{YYYYMMDD} stands for the toolbox revision number.

\bibliographystyle{elsarticle-num} 
\bibliography{./bib/A1_SFP_Toolbox_manually_modified.bib}






\clearpage

\end{document}